# IMAGE SUPER-RESOLUTION VIA DUAL-DICTIONARY LEARNING AND SPARSE REPRESENTATION

*Jian Zhang[a], Chen Zhao[b], Ruiqin Xiong[b], Siwei Ma[b], Debin Zhao[a]*

[a]School of Computer Science and Technology, Harbin Institute of Technology, Harbin, 150001, China
[b]Institute of Digital Media, Peking University, Beijing, 100871, China

## ABSTRACT

Learning-based image super-resolution aims to reconstruct high-frequency (HF) details from the prior model trained by a set of high- and low-resolution image patches. In this paper, HF to be estimated is considered as a combination of two components: main high-frequency (MHF) and residual high-frequency (RHF), and we propose a novel image super-resolution method via dual-dictionary learning and sparse representation, which consists of the main dictionary learning and the residual dictionary learning, to recover MHF and RHF respectively. Extensive experimental results on test images validate that by employing the proposed two-layer progressive scheme, more image details can be recovered and much better results can be achieved than the state-of-the-art algorithms in terms of both PSNR and visual perception.

## 1. INTRODUCTION

Image super-resolution deals with the problem of reconstructing a high-resolution (HR) image from one or several of its low-resolution (LR) counterparts. In this paper, we focus on single image super-resolution, which can be formulated as follows:

$$\mathbf{y} = DH\mathbf{x} + \mathbf{n} \qquad (1)$$

where $H$ denotes a blurring operator, $D$ denotes a decimation operator by a factor $s$, $\mathbf{x}$ is the original HR image, $\mathbf{y}$ is the observed LR image and $\mathbf{n}$ represents an additive i.i.d. white Gaussian noise.

In recent years, example-based super-resolution methods have been an active research topic to recover high-frequency details with the help of a database consisting of co-occurrence examples from a training set of HR and LR image patches [1, 2, 3, 4]. However, the aforementioned example-based algorithms heavily rely on the similarity between the training set and the test set.

Recent achievements in sparse representation suggest that the linear relationships among high-dimension signals can be accurately recovered from their low-dimension projections [5]. Instead of working directly with the patch pairs sampled from high- and low-resolution images, based on the assumption that for a given low-resolution patch, the sparse representation vector in the over-complete dictionary trained by low-resolution images is the same as the one of its corresponding high-resolution patch in the over-complete dictionary trained by high-resolution images, [6] leant a compact representation for these patch pairs to capture the co-occurrence prior to improve the speed and the robustness significantly, achieving state-of-the-art performance. Lately, embarking from the algorithm by [6], a modified approach is proposed [7], which shows to be more efficient and much faster than [6].

However, due to the restriction of the over-complete dictionary's size and the intrinsic sparsity of the algorithm, one limitation in recovering high-frequency details should be noticed. It's difficult to recover the details of the high-frequency in the corresponding HR image completely from the initial interpolation of an input LR image, for the gap between the frequency spectrum of the corresponding HR image and that of the initial interpolation is so wide that learning-based algorithm usually can't work well.

To address above problems, in this paper, the high-frequency (HF) to be estimated is considered as a combination of two components: main high-frequency (MHF) and residual high-frequency (RHF), and we develop a novel method for learning-based image super-resolution via sparse representation, which consists of dual-dictionary learning levels: main dictionary learning and residual dictionary learning, corresponding to recovering MHF and RHF, respectively. Through the proposed two-layer algorithm, in which details of high-frequency are estimated by a progressive way, the main high-frequency is first recovered to reduce the gap of the frequency spectrum primarily, and then the residual high-frequency is reconstructed to enhance the result effectively with a shorter gap of the frequency spectrum. So our scheme can be seen as a coarse-to-fine recovering process and better results can be expected, which will be demonstrated by experimental section hereinafter.

This paper is organized as follows. Section 2 describes details of the proposed super-resolution scheme including dictionary learning stage and image synthesis stage. Experimental results are shown in Section 3, and conclusions are drawn in Section 4.

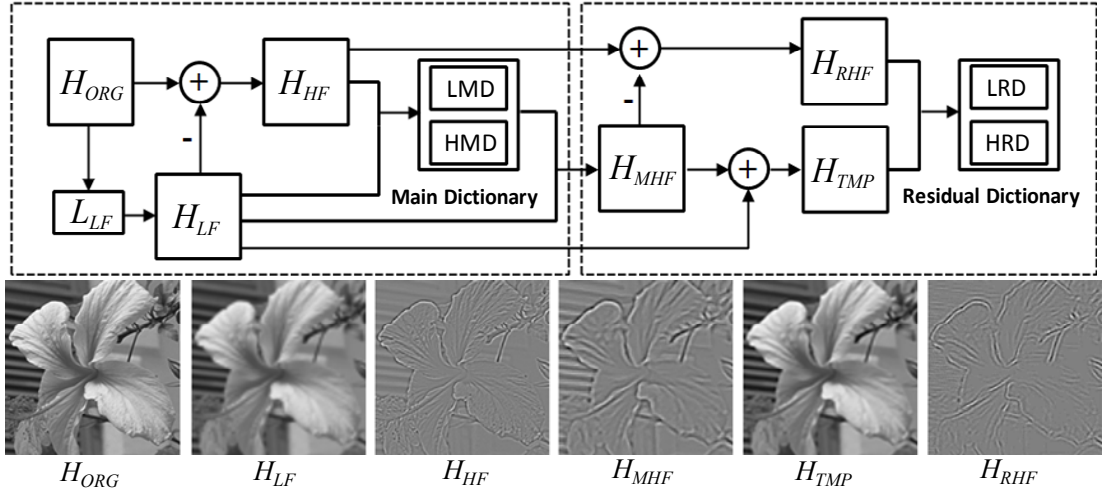

| $H_{ORG}$ | $H_{LF}$ | $H_{HF}$ | $H_{MHF}$ | $H_{TMP}$ | $H_{RHF}$ |

Fig. 1. Illustration of dictionary learning stage.

## 2. THE PROPOSED IMAGE SUPER-RESOLUTION SCHEME

The proposed scheme consists of a dictionary learning stage that trains dual dictionaries, namely main dictionary (MD) and residual dictionary (RD), as described in Figure 1 and an image synthesis stage, performing the image super-resolution on the input image using the trained model from the previous stage, as described in Figure 2.

### 2.1. Dictionary Learning Stage

In this stage, two dictionaries are trained using sparse representation, i.e. MD and RD, which correspond to the recovery of MHF and RHF, respectively. Many methods have been proposed for dictionary learning by sparse representation, such as [8, 9]. In this paper, we adopt the method recently published in [7] for over-complete dictionary learning. Assuming a local Sparse-Land model on image patches, serving as regularization, [7] has shown to have more effectivity and higher efficiency than [6]. More details can be found in [7].

The dictionary learning stage starts by collecting a set of training HR images. As illustrated in Fig. 1, a HR training image denoted by $H_{ORG}$, is first blurred and downsampled to yield a corresponding LR low-frequency image denoted by $L_{LF}$. By applying bicubic interpolation on $L_{LF}$, a HR low-frequency image is constructed, denoted by $H_{LF}$ in Fig.1, which is of the same size as $H_{ORG}$. Then, real HR high-frequency image $H_{HF}$ is generated by subtracting $H_{LF}$ from $H_{ORG}$.

Then, MD will be built, which is actually a combination of two coupled sub-dictionaries: low-frequency main dictionary (LMD) and high-frequency main dictionary (HMD). With $H_{LF}$ and $H_{HF}$, local patches are extracted forming the training data $TD = \{\mathbf{p}_h^k, \mathbf{p}_l^k\}_k$. $\mathbf{p}_h^k$ is the set of patches extracted from the high-resolution image $H_{HF}$ directly, and $\mathbf{p}_l^k$ mean those patches built by first extracting patches from filtered images obtained by filtering $H_{LF}$ with certain high-pass filters such as Laplacian high-pass filters, and then reducing the dimensions by Principal Component Analysis (PCA) algorithm.

Next, the K-SVD dictionary training [8] is applied to the set of patches $\{\mathbf{p}_l^k\}_k$, generating LMD:

$$\mathbf{LMD}, \{\mathbf{q}^k\} = \arg\min_{\mathbf{LMD},\{\mathbf{q}^k\}} \sum_k \left\| \mathbf{p}_l^k - \mathbf{LMD} \cdot \mathbf{q}^k \right\|_2^2 \quad (2)$$
$$s.t. \quad \left\| \mathbf{q}^k \right\|_0 \leq L \quad \forall k,$$

where $\{\mathbf{q}^k\}_k$ are sparse representation vectors, and $\|\cdot\|_0$ is the $l^0$ norm counting the nonzero entries of a vector. Based on the assumption that the patch $\mathbf{p}_h^k$ can be recovered by approximation as $\mathbf{p}_h^k \approx \mathbf{HMD} \cdot \mathbf{q}^k$, HMD can be defined by minimizing the following mean approximation error, i.e.,

$$\mathbf{HMD} = \arg\min_{\mathbf{HMD}} \sum_k \left\| \mathbf{p}_h^k - \mathbf{HMD} \cdot \mathbf{q}^k \right\|_2^2$$
$$= \arg\min_{\mathbf{HMD}} \sum_k \left\| \mathbf{P}_h - \mathbf{HMD} \cdot \mathbf{Q} \right\|_2^2, \quad (3)$$

where the matrices $\mathbf{P}_h$ and $\mathbf{Q}$ consist of $\{\mathbf{p}_h^k\}_k$ and $\{\mathbf{q}^k\}_k$, respectively. Therefore, the solution can be solved as follows (given that $\mathbf{Q}$ has full row rank):

$$\mathbf{HMD} = \mathbf{P}_h \mathbf{Q}^+ = \mathbf{P}_h \mathbf{Q}^T (\mathbf{Q}\mathbf{Q}^T)^{-1}. \quad (4)$$

Finally, the residual dictionary will be trained in the following steps. With the main dictionary and $H_{LF}$, the HR main high-frequency image denoted by $H_{MHF}$ is produced by virtue of image reconstruction method which will be introduced in the next subsection. Utilizing $H_{MHF}$, the HR temporary image denoted by $H_{TMP}$ which contains more details than $H_{LF}$ and the HR residual high-frequency image denoted by $H_{RHF}$ are generated, as shown in Fig. 1. Thus, RD can be built with the input of $H_{TMP}$ and $H_{RHF}$ using the same dictionary learning method as MD.

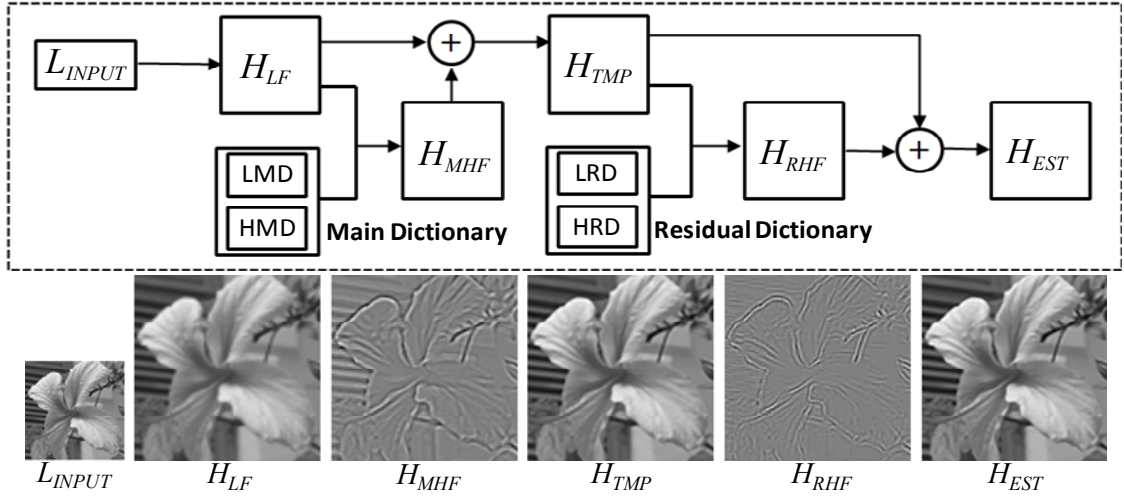

$L_{INPUT}$    $H_{LF}$    $H_{MHF}$    $H_{TMP}$    $H_{RHF}$    $H_{EST}$

Fig. 2. Illustration of image synthesis stage.

It is important to note that RD also consists of two coupled sub-dictionaries: low-frequency residual dictionary (LRD) and high-frequency residual dictionary (HRD), and both MD and RD make up the dual-dictionaries.

**2.2. Image Synthesis Stage**

Image synthesis stage attempts to magnify an input LR image, which is assumed to be generated from a HR image by the same blur and downsample operations as used in the above learning stage. The final estimated HR image is reconstructed by using dual dictionaries successively and more high-frequency details are added progressively, as illustrated in Fig. 2.

To begin with, an input LR image denoted by $L_{INPUT}$ is interpolated by bicubic method to produce a HR low-frequency image denoted by $H_{LF}$. Combining $H_{LF}$ and MD, the HR main high-frequency image denoted by $H_{MHF}$ is generated employing the image reconstruction method in [7]. Concretely, $H_{LF}$ is filtered with the same high-pass filters and PCA projection as the training stage, and then is decomposed into overlapped patches $\{\mathbf{p}_l^k\}_k$. The OMP algorithm [8] is applied to generate $\{\mathbf{p}_l^k\}_k$, and the sparse representation vectors $\{\mathbf{q}^k\}_k$ is built by allocating L atoms to their representation. The representation vectors $\{\mathbf{q}^k\}_k$ are multiplied by HMD, reconstructing high-resolution patches by $\{\hat{\mathbf{p}}_h^k\}_k = \{\mathbf{HMD} \cdot \mathbf{q}^k\}_k$. Defining $\mathbf{R}_k$ the operator which extracts a patch from the high resolution image in location $k$. The HR main high-frequency image denoted by $H_{MHF}$ is generated by solving the following minimization problem:

$$H_{MHF} = \arg\min_{H_{MHF}} \sum_k \left\| \mathbf{R}_k H_{MHF} - \hat{\mathbf{p}}_h^k \right\|_2^2, \quad (5)$$

which has a closed-form Least-Square solution, given by

$$H_{MHF} = [\sum_k \mathbf{R}_k^T \mathbf{R}_k]^{-1} \sum_k \mathbf{R}_k^T \hat{\mathbf{p}}_h^k. \quad (6)$$

Then, the HR temporary image denoted by $H_{TMP}$ containing more details than $H_{LF}$ is built by adding $H_{LF}$ to $H_{MHF}$. Next, by making use of $H_{TMP}$ and RD, the same image reconstruction method is performed again to synthesize the HR residual high-frequency image denoted by $H_{RHF}$. Finally, the HR estimated image denoted by $H_{EST}$ is reconstructed by summing $H_{TMP}$ and $H_{RHF}$, as shown in Fig. 2.

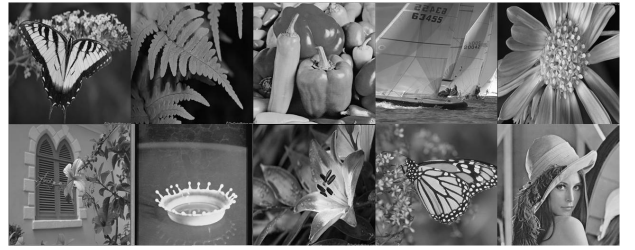

Fig. 3. Dictionary training and performance test images.

## 3. EXPERIMENTAL RESULTS

In this section, we implement our method on some test images to compare with other interpolation methods such as bicubic interpolation method and sparse representation method [7]. The blurring operator is 5×5 Gaussian filter with standard deviation of 1, and the decimation operator is direct downsampling with scale factor 2.

In this paper, the size of MD and the size of RD are both set to 500, while the size of dictionary in [7] is set to 1000, equaling to the sum of the size of MD and the size of RD in the proposed scheme. The number of atoms for representing each image patch is fixed to 3, and the size of image patch is 9×9 with overlap of 1 pixel. The experimental images are shown in Fig. 3, in which the first image is exploited for dictionary learning and the others for performance test. The PSNR comparisons with different methods are listed in Table 1. It can be seen from Table 1 that the proposed method achieves the highest PSNR for all instances, and can improve by 3.14 dB and 0.5 dB with respect to average PSNR compared to Bicubic and Zeyde et. al. [7], respectively. The last column in Table 1 represents the gain obtained by the proposed method over state-of-the-art method [7], validating the necessity and effectivity of dual-dictionary learning.

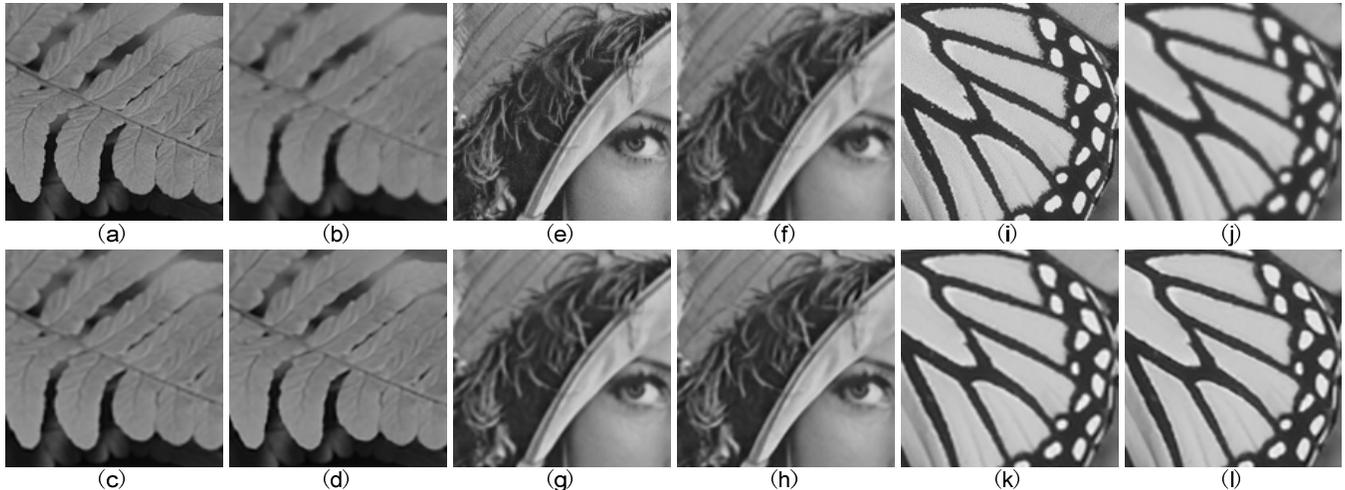

Fig. 4. Visual comparison by different interpolation methods: Portions from three test images. (from left to right: *Foliage, Lena, Monarch*). (a) (e) (i) Original images; (b) (f) (j) Bicubic interpolation; (c) (g) (k) Zeyde et. al. [7]; (d) (h) (l) The proposed algorithm.

Some visual results are shown in Fig. 4. Note that the proposed algorithm performs visually much better than bicubic interpolation, having less visual artifacts and producing sharper results. Compared with [7], the proposed algorithm provides more image details with improved PSNR.

Table 1. PSNR comparisons with different algorithms (dB)

| Images | Bicubic | Zeyde et.al. [7] | Proposed | Gain |
|---|---|---|---|---|
| *Foliage* | 31.65 | 34.73 | 35.50 | **0.77** |
| *Mum* | 31.05 | 34.65 | 35.30 | **0.65** |
| *Monarch* | 27.78 | 30.34 | 30.88 | **0.54** |
| *Peppers* | 32.32 | 34.46 | 34.78 | **0.32** |
| *Flower* | 32.12 | 34.97 | 35.54 | **0.57** |
| *Window* | 31.19 | 33.74 | 34.20 | **0.46** |
| *Sailboat* | 30.56 | 32.36 | 32.80 | **0.44** |
| *Splash* | 36.16 | 39.07 | 39.50 | **0.43** |
| *Lena* | 32.19 | 34.66 | 34.96 | **0.30** |
| *Average* | 31.69 | 34.33 | 34.83 | **0.50** |

## 4. CONCLUSIONS

This paper presents a novel image super-resolution approach via dual-dictionary learning and sparse representation, which can reconstruct lost high-frequency details by a two-layer progressive way utilizing distinct dictionaries. Experimental results show that the proposed method is able to remove some restrictions of frequency spectrum suffered by traditional example-based methods which lead to missing much image details, and achieve better results in terms of both PSNR and visual perception.

## 5. ACKNOWLEDGEMENT

This work was supported in part by National Basic Research Program of China (973 Program, 2009CB320904 and 2009CB320905), National Natural Science Foundation of China (No. 61073083 and No. 60833013), Beijing Natural Science Foundation (No. 4112026) and Specialized Research Fund for the Doctoral Program of Higher Education (No. 20100001120027).